\documentclass[letterpaper, 10 pt, conference]{ieeeconf}

\IEEEoverridecommandlockouts
\usepackage{graphics}
\usepackage{epsfig}
\usepackage{mathptmx}
\usepackage{amsmath}
\usepackage{amssymb}
\usepackage{algorithm}
\usepackage{algpseudocode}
\usepackage{url}
\usepackage{cite}
\usepackage{multirow} 

\usepackage{capt-of}  

\usepackage{censor}

\usepackage{tikz}

\usepackage[T1]{fontenc}

\title{\LARGE \bf
LITHE: Bridging Best-Effort Python and Real-Time C++
\\for Hot-Swapping Robotic Control Laws on Commodity Linux
}

\author{
He Kai Lim$^{1}$, Tyler R. Clites$^{1}$%
\thanks{$^{1}$All authors are with the Department of Mechanical and Aerospace Engineering, University
of California Los Angeles, Los Angeles, CA 90095 USA {\tt\small limhekai@ucla.edu, clites@ucla.edu}}%
}

\begin{document}

\maketitle

\begin{tikzpicture}[remember picture,overlay]
\node[anchor=south, yshift=10pt] at (current page.south) {\fbox{\parbox{\dimexpr\textwidth-\fboxsep-\fboxrule\relax}{\footnotesize This work has been submitted to the IEEE for possible publication. Copyright may be transferred without notice, after which this version may no longer be accessible.}}};
\end{tikzpicture}

\thispagestyle{empty}
\pagestyle{empty}

\begin{abstract}
Modern robotic systems rely on hierarchical control, where a high-level ``Brain" (Python) directs a lower-level ``Spine" (C++ real-time controller). Despite its necessity, this hierarchy makes it difficult for the Brain to completely rewrite the Spine's immutable control logic, and consequently inhibits fundamental adaptation for different tasks and environments. Conventional approaches require complex middleware, proprietary hardware, or sacrifice real-time performance. We present LITHE (Linux Isolated Threading for Hierarchical Execution), a lightweight software architecture that collapses the robot control hierarchy onto a commodity single-board computer (Raspberry Pi 4B with \textit{pi3hat}, 250 USD), while maintaining safe frequency decoupling between the Brain and Spine. LITHE integrates strict CPU isolation (\texttt{isolcpus}), lock-free inter-process communication (IPC), and pipelined execution to achieve functionally real-time code execution that meets high-frequency deadlines with minimal jitter. By further adding multi-threaded dynamic linking, LITHE enables a Python-based Brain to govern and dynamically evolve the logic of a 1kHz real-time C++ Spine without interruption. We validate that the system achieves ``functional real-time" performance with worst-case execution time (WCET) $<100\,\mu$s and maximum release jitter (MRJ) $<4\,\mu$s under heavy computational load. We also demonstrate a novel application where a decentralized large language model (LLM) supervisor (or any other complex model) can perform online system identification and evolve a real-time controller on-the-fly, without interrupting the 1 kHz control loop. In essence, LITHE eliminates the ``immutable compiled code" bottleneck for real-time control by allowing best effort Brains to safely synthesize and inject completely new control laws into an active real-time Spine. Ultimately, this bridges a critical gap between high-level artificial intelligence and low-level real-time control, thereby unlocking new capabilities in safe, human-in-the-loop robotic systems via continuous real-time evolution of its embodied intelligence.

The LITHE codebase will be open-sourced upon acceptance. 
\end{abstract}

\section{Introduction}
\label{sec:introduction}

\begin{figure}[t]
  \centering
  \includegraphics[width=1\linewidth]{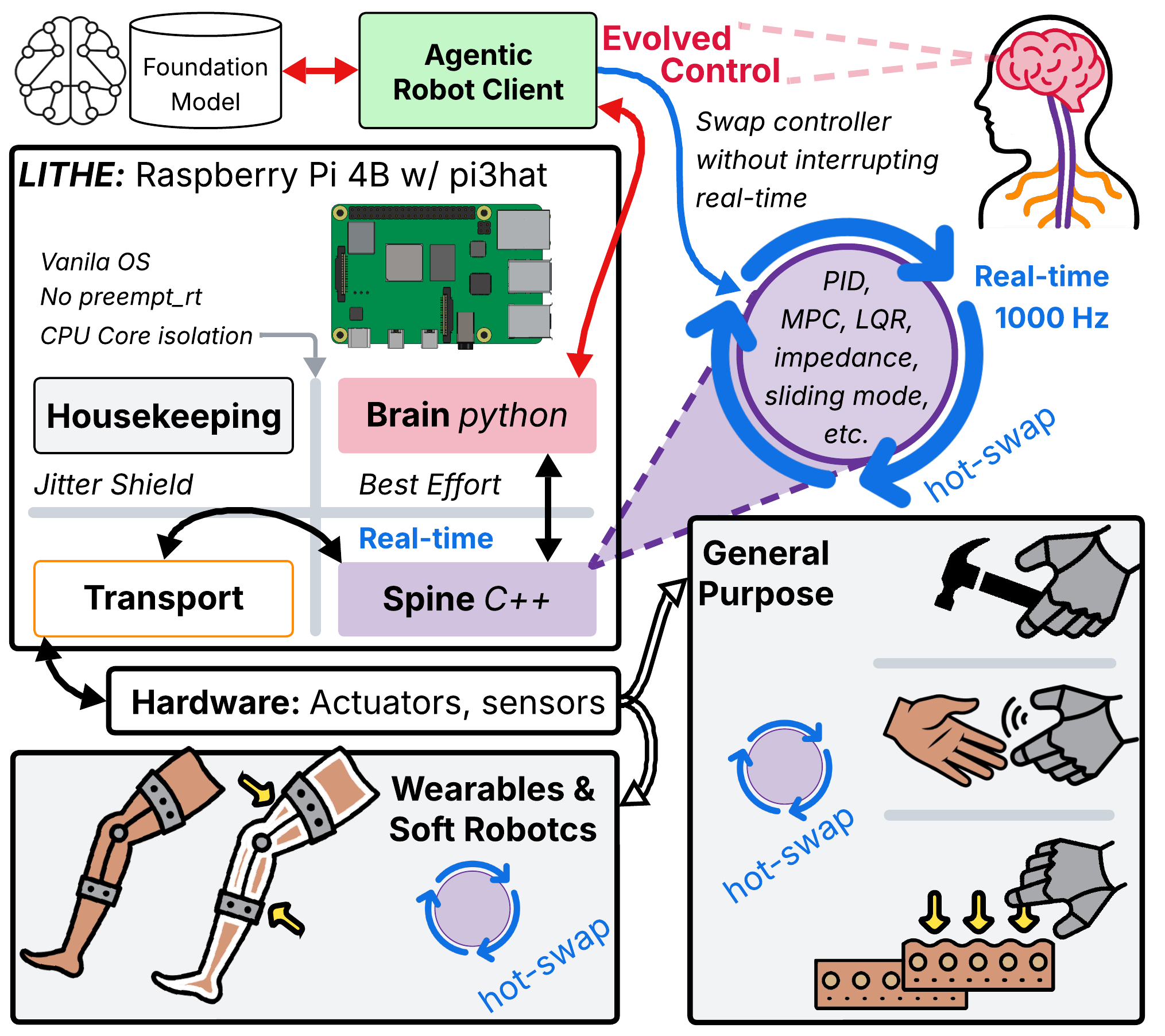}
  \caption{\textbf{Hierarchical control from the Brain, across the Spine, onto different tasks and applications.} LITHE enables high-level ``best-effort" models and algorithms to evolve and hot-swap low-level real-time controllers without interrupting on-the-fly real-time operation. This empowers continuous, real-time evolution of embodied intelligence.}
  \label{fig:teaser}
\end{figure}

Robot control is typically hierarchical\cite{merelHierarchicalMotorControl2019, saridisIntelligentRoboticControl1983, haighHighlevelPlanningLowlevel1997}. At the top, a ``Brain'' performs high-level decision-making, relying on machine learning (ML) algorithms within the Python ecosystem (e.g., PyTorch) \cite{merelHierarchicalMotorControl2019, rybczakSurveyMachineLearning2024, aljamalComprehensiveReviewRobotics2025, wangPyPoseLibraryRobot2023, delhaissePyRoboLearnPythonFramework2020}. These tools offer immense feature richness, but require a full operating system (OS) due to their complex dependencies and managed runtime. Below the Brain, a ``Spine'' handles the immediate high-frequency command of robotic hardware such as motor control and sensor sampling \cite{merelHierarchicalMotorControl2019, saridisIntelligentRoboticControl1983}. To ensure safety and determinism, the Spine is typically developed in C++ and deployed as firmware on embedded microcontrollers or FPGAs \cite{barrProgrammingEmbeddedSystems1999}.

While this hierarchy is necessary for performance, it introduces a bottleneck for general-purpose robotics: immutable compiled code \cite{marevacFrameworkDesignDynamic2025, hoplerVersatileToolboxModel2001, ichnowskiMotionPlanningTemplates2019}. During robot operation, structural changes to real-time control logic (beyond parametrization) are not easily redeployed on embedded hardware \cite{marevacFrameworkDesignDynamic2025, hoplerVersatileToolboxModel2001, ichnowskiMotionPlanningTemplates2019}. In most architectures, robot operation must be halted so that new C++ binaries can be flashed and reloaded. The immmutable nature of real-time logic thus inhibits dynamic general-purpose utility of robots in the field, where pausing operation may not be possible or desirable \cite{yaacoubSchedulingDynamicSoftware2023}.

This immutability is also detrimental to modern paradigms like Reinforcement Learning \cite{wangPyPoseLibraryRobot2023, paneReinforcementLearningBased2019} and Morphological Evolution \cite{xinDirectedMorphologicalEvolution2026}, where the controller must evolve in tandem with the learned action itself \cite{yaacoubSchedulingDynamicSoftware2023, paneReinforcementLearningBased2019, xinDirectedMorphologicalEvolution2026}. We illustrate these limitations in the field of wearable soft robotics and prosthetics. Here, the robot controller must evolve and adapt to viscoelasticity and creep of soft biological tissue over hours of wear, across a spectrum of predicted and \textit{unpredicted} activities, while preserving safe and stiff real-time operation for the user. In these human-in-the-loop applications, the immutability of compiled real-time code is not merely inconvenient; it is a barrier to continuous learning and evolution of safe embodied intelligence.

Current approaches to tackle the immutable real-time problem force a compromise between accessibility and performance. Middleware approaches add interoperability but introduce real-time complexity \cite{mohamedMiddlewareRoboticsSurvey2008, gamboSystematicLiteratureReview2025}. State machines can be built entirely with bare-metal C++ firmware, but lack flexibility for adaptation beyond the predicted states or predetermined architecture. Python's runtime interpretation promotes maximum flexibility, albeit with relinquished real-time response \cite{choFeasibilityStudyPythonBased2023}. Commercial real-time prototyping systems address these issues but can be prohibitively expensive and closed-source. To explore paradigms for learning and evolving real-time control, new architectures are needed that combine flexible interpreted languages with functionally ``good enough" real-time execution on accessible hardware.

In this work, we present LITHE (Linux Isolated Threading for Hierarchical Execution), a system designed to accessibly bridge control hierarchies using low-cost hardware. We collapse the Brain-Spine interface on a commodity computer (Raspberry Pi 4B with \textit{pi3hat}, 250 USD) without sacrificing functional real-time performance. Our contributions are:

\begin{enumerate}
    \item \textbf{Frequency-Decoupled Architecture:} We demonstrate that a vanilla Linux kernel (without \texttt{PREEMPT\_RT}), when configured with strict CPU isolation, can sustain a functionally real-time C++ Spine (1 kHz) on one core while simultaneously running a best effort Python Brain on another. This ``user-space real-time" environment ensures the Brain's jitter and latency do not affect the Spine's real-time performance.
    
    \item \textbf{Hot-Swap Loader:} 
    We introduce a mechanism for end-to-end injection of new real-time control laws into an active real-time system. It uses careful core allocation to invoke \texttt{gcc}, multi-threaded dynamic linking (\texttt{dlopen}), and atomic pointer swaps to update the active Spine thread. This empowers unlimited evolution of the control law across an infinite solution space, without missing or breaking real-time requirements.
    
    \item \textbf{Benchmarks and Demonstration:} We benchmark the architecture through standard stress-tests, and we demonstrate its viability for dynamically-evolved real-time control. A simple agentic robot client, based on qwen2.5-coder-7b large language model (LLM), is tasked to perform system identification and evolve a real-time control law to reduce tracking error for a 1-DOF robot. This shows that commodity hardware running our LITHE architecture maintains safety and low jitter even when integrated with a non-time-sensitive, decentralized, model-evolving, robot control system.
\end{enumerate}

The LITHE codebase will be open-sourced upon acceptance. 
Our results demonstrate that bridging hierarchies across the robot control stack can empower high-latency ML models to safely synthesize and hot-swap real-time control laws, even on low-cost commodity hardware. By dynamically connecting the ``thinking" Brain with the ``acting" Spine, LITHE creates a platform for continuous real-time evolution of embodied intelligence.

\begin{figure*}[t]
  \centering
  \includegraphics[width=1\linewidth]{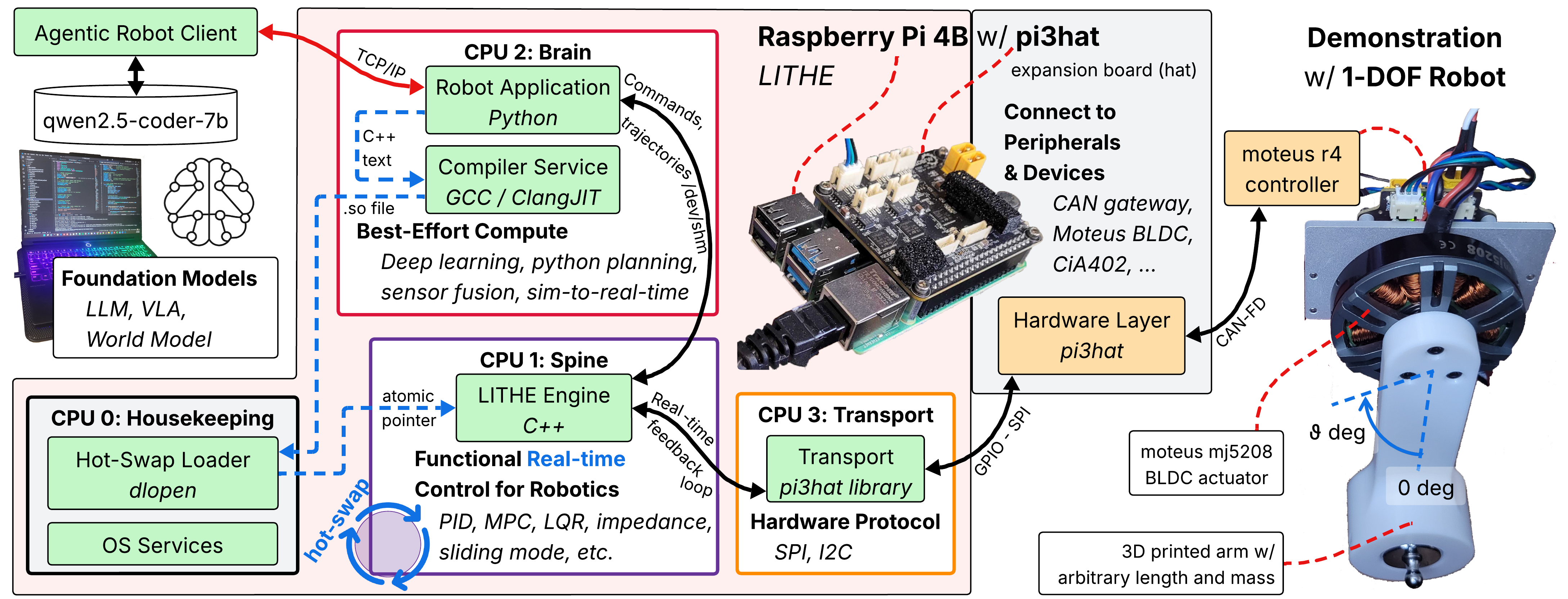}
  \caption{\textbf{LITHE system architecture for evolving real-time robot control on a Raspberry Pi.}
    A top level “Python Brain” is isolated to CPU 2. It can run independently, with best-effort tasks such as deep learning, or in collaboration with an external agentic robot client. The Brain sends commands and trajectories via lock-free inter-process communication to a low-level “C++ Spine” which is isolated on CPU 1. The Spine maintains a functional real-time controller, such as PID or impedance control. Through our hot-swap mechanism, this controller can be hot-swapped on-the-fly, without interrupting real-time. It interacts with robot hardware through an isolated transport layer on CPU 3, based on the open-source pi3hat library for the pi3hat GPIO expansion board.
    For our simple experiment in Section \ref{sec:experiment_supervised}, we set up an external agent based on qwen2.5-coder-7b to collaborate with the Brain. Through LITHE, we then implement various hot-swapped controllers that interact with a 1-DOF robot comprised of moteus hardware and a 3D printed arm.}
  \label{fig:architecture}
\end{figure*}


\section{Related Work}
\label{sec:relatedwork}

\subsection{Architectures for Real-Time Control}
Existing solutions for hierarchical real-time robot control generally fall into two categories: proprietary rapid control prototyping (RCP) systems, and open-source Linux patches. Commercial RCP solutions (e.g., dSPACE, Speedgoat) offer verified real-time performance and ease of use, but are prohibitively expensive and closed-source. This restricts their accessibility for scalable research. Conversely, the Linux \texttt{PREEMPT\_RT} patch offers an open-source alternative, but requires advanced knowledge of kernel compilation, strict constraints on C++ driver development (e.g., avoiding page faults and standard heap allocation), and hardware-specific tuning \cite{maddenChallengesUsingLinux2019}. Consequently, these patches are brittle, often breaking with closed-sourced driver updates (e.g., NVIDIA GPU drivers). Furthermore, Deng et al. \cite{dengInterferencefreeOperatingSystem} demonstrated that a patched kernel still suffers from ``interference" where high-priority real-time threads remain subject to cache pollution from background OS tasks.

At the middleware layer, the state-of-the-art for modular robotics is ROS 2, particularly its real-time variants. While powerful, ROS 2 introduces architectural overhead via the DDS (Data Distribution Service) layer and serialization costs. Achieving true sub-millisecond determinism with ROS 2 remains a subject of active research \cite{gamboSystematicLiteratureReview2025}, often requiring significant tuning. In tandem, Nichols et al. \cite{nicholsRealTimeMotorControl2025} demonstrated the viability of a leaner, Raspberry Pi-based architecture for centralized prosthetic control, distributing tasks across cores with their own, simpler ROS-based software.

Our work builds directly on these foundations but bypasses these software complexities by adopting a ``user-space real-time" strategy to achieve functional real-time, albeit without strictly guaranteed execution. Instead of modifying the scheduler (as in \texttt{PREEMPT\_RT}), we strictly isolate CPU cores to exclude the scheduler entirely. This approach enables standard user-space C++ development without the fragility of custom kernels or the cost of proprietary hardware. Furthermore, we discard the publisher-subscriber model entirely in favor of lock-free, atomic shared memory for minimum latency and maximum flexibility, and we enable device polymorphism by introducing a unified Hardware Abstraction Layer (HAL) to support a ``Frankenstein" mix of actuators (e.g., Moteus, industrial servos).

\subsection{Dynamic Reconfigurability and Hot-Swapping}
Traditional adaptive control strategies rely on mathematical switching (e.g., state machines or control barrier functions) within a single compiled binary. For dynamic reconfiguration of these control laws, modern architectures often use middleware plugins, such as ROS 2's \texttt{pluginlib} \cite{PluginlibROSWiki}, combined with zero-copy IPC transports like Eclipse Iceoryx. These systems still require pre-compiled shared objects and rely on \texttt{dlopen} calls, which introduce non-deterministic blocking and will often lead to missed control deadlines. Further, even if deployed on a \texttt{PREEMPT\_RT} patched kernel, they would be quite complex and require careful management of thread priorities and synchronization to ensure real-time performance.

To our knowledge, LITHE is the first documented architecture that diverges from these works by only using core isolation (\texttt{isolcpus}) to handle just-in-time compilation (\texttt{gcc}) and loading (\texttt{dlopen}) entirely in the background. By further embedding the ability to hot-swap real-time control laws with a wait-free atomic pointer swap on the isolated real-time Spine thread, LITHE safely swaps the \textit{entire binary logic} of a controller inside a 1 kHz loop without dropping a single cycle or losing continuous controller state. This unique capability bridges the gap between high-level ML algorithms (which evolve dynamically) and low-level physical firmware (which is historically immutable), enabling a new class of ``evolving real-time control" via decentralized ML methods.

\section{SYSTEM ARCHITECTURE}
\label{sec:architecture}

\subsection{Commodity Hardware Platform \& Abstraction}
Our hardware selection prioritizes accessibility and open-source maturity. The core compute node is a 250 USD Raspberry Pi 4B with 8GB RAM (RPI) paired with a \textit{pi3hat} expansion board (mjbots Robotic Systems, Cambridge, MA) (Fig. \ref{fig:architecture}). We prioritized the RPI 4B over newer alternatives (RPI 5, NVIDIA Jetson) to minimize cost and maximize compatibility with standard kernels and expansion boards. The \textit{pi3hat} provides five dedicated CAN-FD buses and three onboard STM32 processors to handle low-level signal timing. It offers a high-bandwidth, low-latency link to distributed actuators without the complexity of custom carrier boards.

To ensure our architecture remains hardware-agnostic, we introduce a unified Hardware Abstraction Layer (HAL). The HAL maps diverse hardware protocols from CAN-FD (e.g., Moteus) and CANopen (e.g., CiA 402, Teensy) into a single coherent memory map. This allows construction of ``Frankenstein" systems that mix high-performance brushless motors with commodity sensors (e.g., Teensy 4.1 CAN gateways for sensor sampling) without modifying the core control loop. 

\subsection{Operating System Tuning}
Standard Linux schedulers prioritize throughput over task latency, making them ill-suited for real-time robot control \cite{wongAchievingFairnessLinux2008, loziLinuxSchedulerDecade2016}. Rather than patching the kernel to preempt housekeeping tasks (as in \texttt{PREEMPT\_RT}), we restrict \textit{where} the kernel is allowed to schedule them. We use vanilla Raspberry Pi OS Lite (Kernel 6.12.47) (no \texttt{PREEMPT\_RT}), and implement a strict CPU isolation strategy \cite{dengInterferencefreeOperatingSystem}. By applying a comprehensive sequence of boot parameters (e.g., \texttt{isolcpus=1-3}), system configuration commands for high performance computing (e.g., \texttt{nohz\_full}, \texttt{rcu\_nocbs=1-3}), and extensive IRQ affinity tuning (see LITHE project website for full list), we partition the quad-core processor into four distinct functional domains (Fig. \ref{fig:architecture}):

\begin{itemize}
    \item \textbf{CPU 0 (Housekeeping):} Handles all Linux housekeeping, SSH sessions, and non-critical interrupts (IRQs). It acts as a shield, absorbing most system jitter.
    \item \textbf{CPU 1 (Spine):} Dedicated exclusively to the C++ Control Loop. With interrupts offloaded, this core operates in a ``user-space real-time" mode, running a single \texttt{while} loop with practically no scheduler interference.
    \item \textbf{CPU 2 (Brain):} Hosts the high-level Python runtime. Since the Python Global Interpreter Lock (GIL) is single-threaded, pinning the Python process here ensures it cannot contend with the Spine by default.
    \item \textbf{CPU 3 (Transport):} Manages blocking I/O operations (SPI/CAN interactions via \textit{pi3hat}). This separation ensures that waiting for bus traffic never stalls the control logic on CPU 1. This process is implemented entirely using the open-source \textit{pi3hat} library \cite{MjbotsPi3hat2026}.
\end{itemize}

\subsection{Inter-Process Communication (IPC)}
IPC strategies are fundamental to robotic control because they enable information exchange between different hardware, programs, threads, and cores in the robot system \cite{wangTZCEfficientInterProcess2019}. To bridge the non-time-sensitive Python Brain and the real-time C++ Spine, we employ a lock-free, zero-copy IPC mechanism based on POSIX Shared Memory (\texttt{/dev/shm}). To ensure data consistency across the language barrier, we utilize a custom schema-driven code generator. This build-time tool defines the memory layout for all robot states and commands, generating both the C++ structs and corresponding Python \texttt{ctypes} bindings. It strictly enforces ABI compatibility and byte alignment, eliminating serialization overhead and protecting against misalignment bugs.

At runtime, data access is managed via a Seqlock (Sequence Lock) pattern \cite{lameterEffectiveSynchronizationLinux20}. This uses atomic sequence counters to ensure consistency: if the Spine attempts to read a command while the Brain is writing, the sequence check fails and the read is retried. This non-blocking design allows the Python Brain to pause for garbage collection without ever stalling the real-time Spine loop. To mitigate the frequency mismatch between the nondeterministic Brain (typically 50--100 Hz) and the 1 kHz Spine, we implement a Catmull--Rom Spline interpolator in the Spine's command handler. The Brain publishes a trajectory of future waypoints, and the Spine smoothly interpolates between them to maintain $C^1$ continuity at the actuator level.

\subsection{Spine Pipelined Execution}
\begin{figure}[t]
  \centering
  \includegraphics[width=1.0\linewidth]{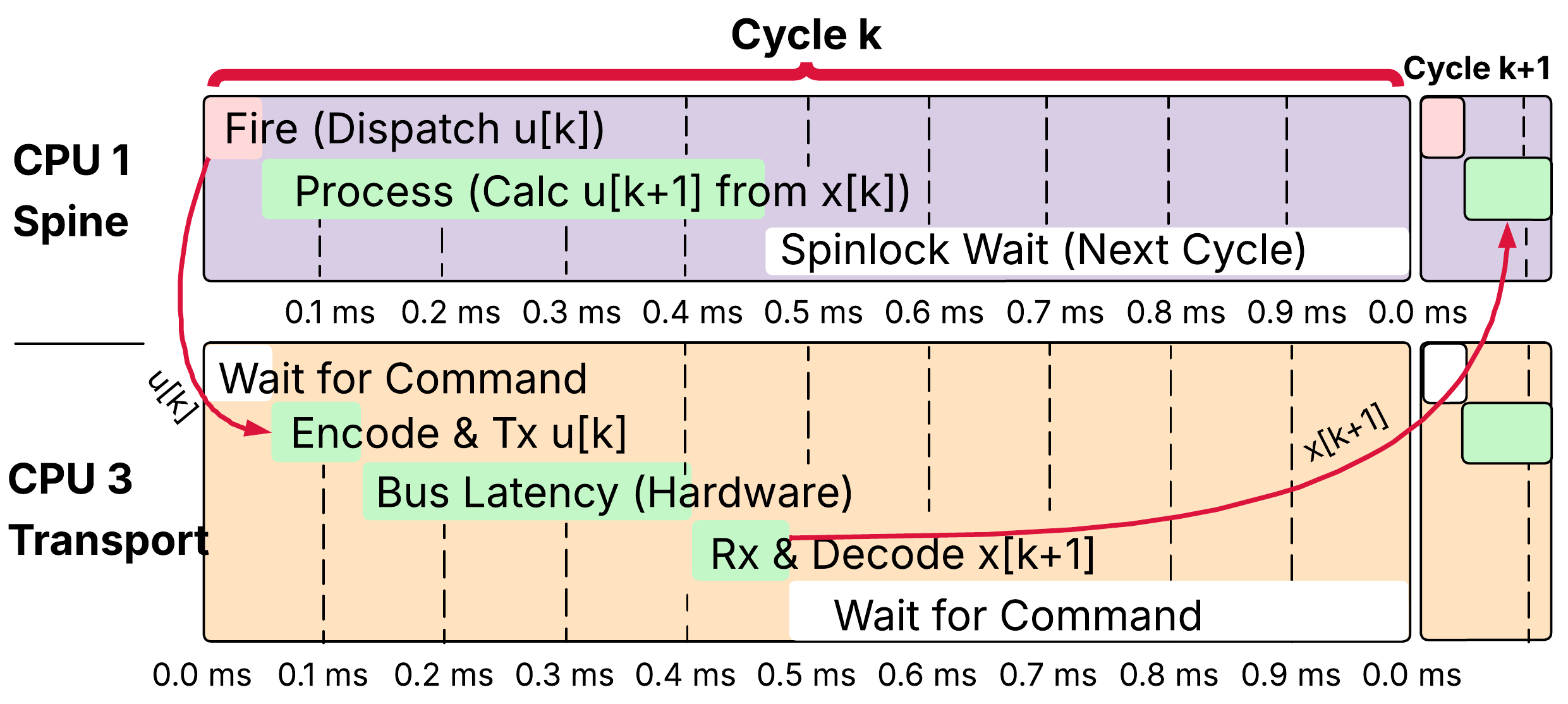}
  \caption{\textbf{Example timing diagram of the pipelined (double buffered) execution for Cycle $k$}. Spine dispatches the command $u_k$ (computed in the previous cycle) and immediately begins computing the next control law $u_{k+1}$ while Transport handles the blocking CAN-FD bus transactions, before returning feedback $x_{k+1}$ for the next cycle. This parallel structure allows the control logic to execute during the hardware transmission delay.}
  \label{fig:pipeline}
\end{figure}

Simple control loops are often blocking: \textit{Read $\to$ Compute $\to$ Write}. This leaves the CPU idle during bus transactions, wasting valuable compute time. Our system masks communication latency by offloading bus transactions to a separate core (Fig. \ref{fig:architecture}, Fig. \ref{fig:pipeline}). Here, we maximize determinism within the Spine by adopting a pipelined (i.e. double buffered) architecture (Fig. \ref{fig:pipeline}) where the system multi-threads with the Transport to fire the previously computed command, and simultaneously processes the next control law. 

In cycle $k$, the Spine transmits command $u_k$ (computed in previous cycle) and immediately begins computing $u_{k+1}$ using sensor data $x_k$ (received from previous cycle). This architecture introduces a deterministic one-cycle delay ($Z^{-1}$) but ensures that the Spine is continuously computing control laws while the Transport manages the bus. To enforce timing, the Spine thread utilizes a busy-wait spinlock on an atomic flag. This keeps the CPU cache hot and prevents core deprioritization. We chose this approach for the tightest theoretical timing precision over imprecise \texttt{sleep()} calls, though async \texttt{await} methods may also be acceptable. The system reliably sustains a 1 kHz loop rate, accommodating the worst-case round-trip latency of the CAN-FD bus over \textit{pi3hat}, and is far beyond any computational bottleneck of LITHE (Section \ref{disc:limits}). If any timing violations occur (e.g., bus is delayed, seqlock unresolved) the system is configured to either skip a cycle or engage a safety stop; however, our experimental validation (Section \ref{sec:experiments}) confirms that no deadline misses occur at the nominal 1 kHz rate.

\subsection{Just-In-Time Hot-Swapping}
LITHE uses \texttt{dlopen} dynamic linking to hot-swap control logic without breaking the real-time loop. To address jitter in system calls from standard dynamic linking (e.g., disk I/O, symbol resolution, memory allocation), we strictly decouple \textit{loading} from \textit{activation} (Fig. \ref{fig:architecture}), while locking process memory into RAM with \texttt{mlockall(MCL\_CURRENT | MCL\_FUTURE)} to prevent page faults.

\subsubsection{The Loader Thread (CPU 0)}
When the Brain requests a new controller, the heavy lifting is delegated to a ``loader" thread pinned to the housekeeping core. This thread handles file system access, loads the shared object (\texttt{.so}) into memory, and resolves function symbols. Because it runs on CPU 0, any page faults or cache pollution are isolated, physically separated from the critical control path on the CPU 1 (Spine).

\subsubsection{The Atomic Handover (CPU 1)}
Once the new function pointer is resolved, the Loader sets an atomic \texttt{Ready} flag. The Spine thread checks this flag at the end of its execution cycle. If set, it performs an atomic pointer swap. This operation is a single CPU instruction, ensuring the handover introduces zero jitter. The old controller is then passed back to the housekeeping thread for safe unloading (\texttt{dlclose}), ensuring that memory deallocation never blocks the real-time loop. Importantly, to prevent output discontinuities (e.g., integrator windup) during the swap, we design control laws to maintain their internal state variables (e.g., integral errors, filtered velocities) within the persistent shared memory block, rather than within the C++ object instance.

\section{EXPERIMENTS \& EVALUATION}
\label{sec:experiments}

We evaluated the LITHE architecture on two fronts: (1) a benchmark of real-time isolation under heavy computational stress, and (2) a practical demonstration of dynamic real-time hot-swapping on a physical robot, as orchestrated by a simulated decentralized robot agent.

\subsection{Real-Time Isolation Benchmark}
To rigorously quantify the efficacy of our CPU isolation, we subjected the Brain to standard stress tests designed to flood the system's shared resources while measuring temporal determinism of the Spine and Transport.

\begin{figure}[t]
  \centering
  \includegraphics[width=1.0\linewidth]{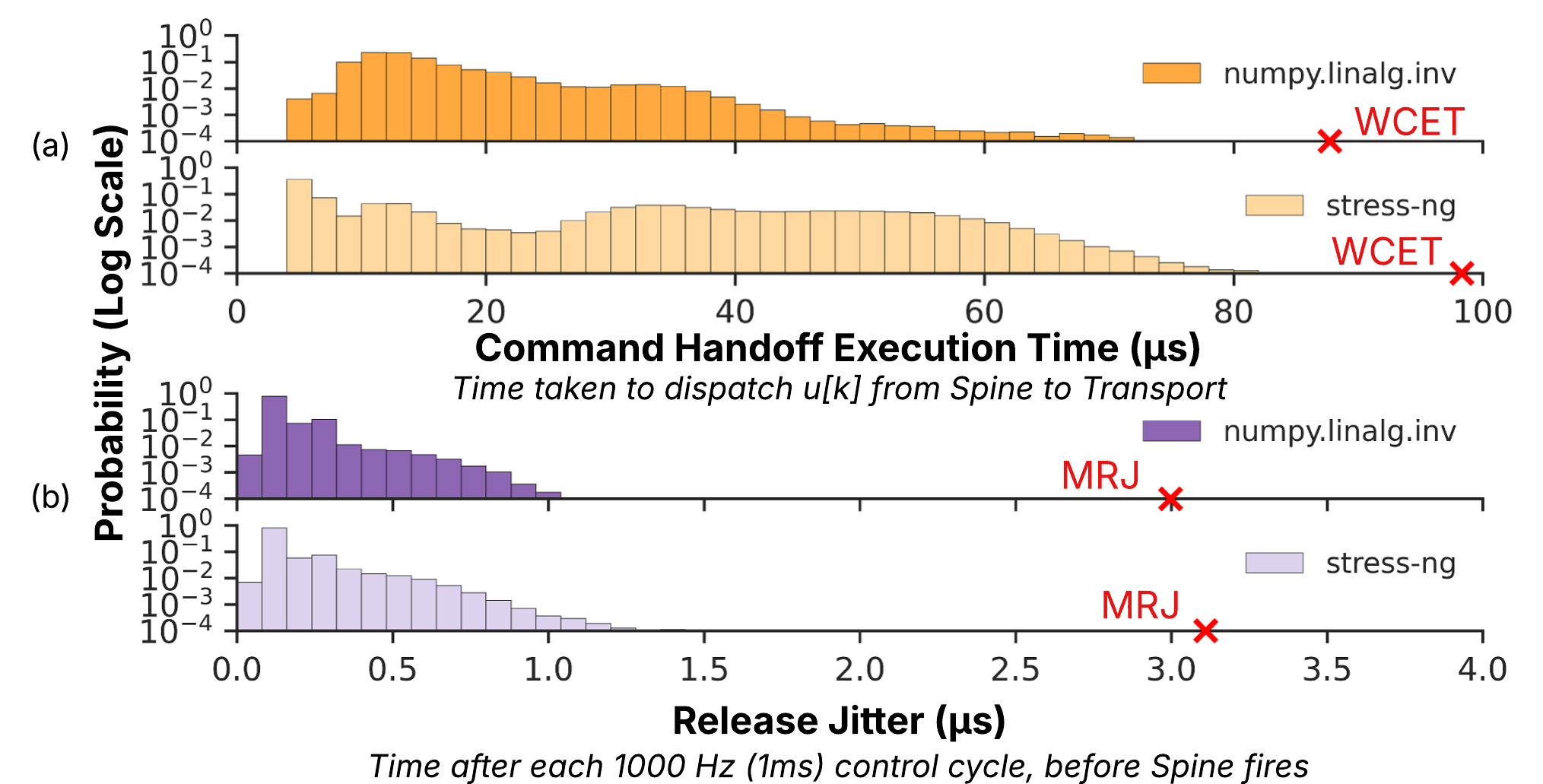}
  \caption{\textbf{Jitter distributions of the Spine under heavy synthetic load on the isolated Brain.} (a) Execution Times for Spine to dispatch command u[k] under heavy synthetic load (NumPy, Stress-ng). Range is [0\,$\mu$s, 100\,$\mu$s], with 2\,$\mu$s bin size. The worst case execution time (WCET) of transport handoff under NumPy is 87.704\,$\mu$s, and for Stress-ng is 98.299\,$\mu$s. (b) Release Jitter Distributions under heavy synthetic load (NumPy, Stress-ng). Range is [0\,$\mu$s, 4\,$\mu$s], with 0.08\,$\mu$s bin size. The maximum release jitter (MRJ) for NumPy is 2.997\,$\mu$s, and for Stress-ng is 3.110\,$\mu$s.}
  \label{fig:jitter}
\end{figure}

\subsubsection{Setup}
The Spine ran a 1 kHz control loop, logging its timing with nanosecond precision using \texttt{std::chrono}. Simultaneously with this operation, the Brain was subjected to two distinct stress workloads:
\begin{itemize}
    \item \textbf{Cache Thrashing (\texttt{stress-ng}):} The standard \texttt{stress-ng --cpu-method fft --vm 1 --vm-bytes 64M} \cite{kingStressng}. This performs Fast Fourier Transforms with aggressive memory allocation, which saturates the L2/L3 caches and memory bus.
    \item \textbf{Thread Sprawl (NumPy):} A continuous loop of random $600 \times 600$ matrix inversions using \texttt{numpy.linalg.inv}. NumPy releases the Python GIL for linear algebra operations and spawns a thread pool that aggressively attempts to utilize all available system cores. This tests LITHE's ability to repel ``rogue" threads from the isolated Spine core.
\end{itemize}

\subsubsection{Metrics}
We tracked two key metrics over a 5-minute duration (300,000 cycles):
\begin{itemize}
    \item \textbf{Command Handoff Execution Time:} The variability in time required to dispatch commands u[k] (Fig. \ref{fig:pipeline}). This metric captures the system's susceptibility to bus contention and memory bandwidth saturation, which are on the critical path for physical actuation.
    \item \textbf{Release Jitter:} The error between the Spine's scheduled fire time and the actual fire time (Fig. \ref{fig:pipeline}). This measures the OS scheduler's ability to keep real-time promises in sending commands to its mechatronic hardware.
\end{itemize}

\subsubsection{Results}
As shown in Figure \ref{fig:jitter}a, the worst-case execution time (WCET) plateaued at 98.3\,$\mu$s under maximum load. Given a 1000\,$\mu$s control period, this preserves a safety margin of $\approx10\times$, confirming that the system sustains 1 kHz operation even when the Brain is maximally loaded. Separately, Figure \ref{fig:jitter}b demonstrates that release jitter remained negligible, never exceeding 3.11\,$\mu$s across both stress conditions, with most cycles falling below 0.5\,$\mu$s.

\subsection{Supervised Evolution of Real-time Control}
\label{sec:experiment_supervised}
To assess LITHE's ability to bridge high-level ML with low-level real-time control, we let a decentralized supervisory agent probe and evolve a real-time control law for a 1-DOF robot arm (Fig. \ref{fig:architecture}), to reduce tracking error.

\subsubsection{Setup} 
A 3D-printed arm (PET-G material) (A1, Bambu Lab, Shenzhen, China) with an unmodeled payload was coupled to a Moteus mj5208 actuator. LITHE interfaced with the motor at 1 kHz. An external workstation (Intel i9-14900HX 32GB, NVIDIA RTX 4070) was set up to host a local large language model (LLM) (qwen2.5-coder-7b) acting as a Supervisory Agent (Fig. \ref{fig:architecture}). This agent parsed telemetry from LITHE and generated C++ source code to evolve the robot's real-time controller for different goals.

We utilized an LLM as a proxy for best effort, high-latency machine-learning workloads. Our goal is to demonstrate that LITHE can execute hot-swaps generated by an external ``black box agent" without violating strict real-time constraints with commodity mechatronics. To turn the LLM into an agent for our experiment, our prompts described the physical context of the 1-DOF robot. Wrappers and templates around the LLM's text response were used to parse telemetry data, and for error-handling by extracting specific snippets of C++ code text that could be sent to LITHE for hot-swapping (see project website for details).

\subsubsection{Experimental Protocol} 
\begin{figure}[t]
  \centering
  \includegraphics[width=\linewidth]{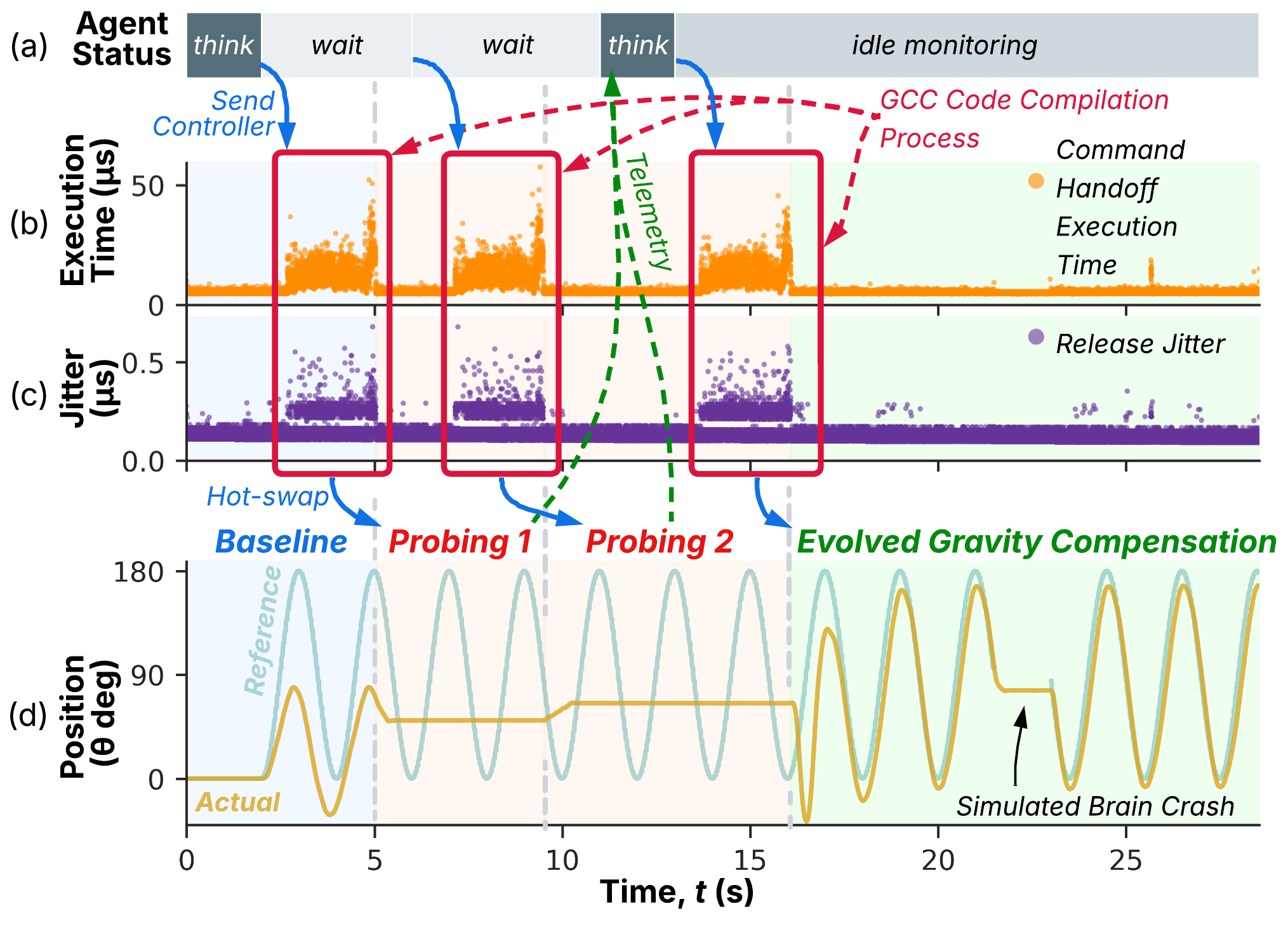}
  \caption{\textbf{Dynamic evolution of real-time control law for robot arm through an external decentralized robot agent.} The agent is qwen2.5-coder-7b, a large language model (LLM)
  (a) The agent starts by thinking about our prompts of the system. It formulates probing controllers for system identification, and a final gravity compensation controller based on received telemetry (green arrows). These controllers are sent at approximately $t=2, 6, 13$s (blue arrows), and correspond with increased LITHE jitter due to subsequent compilation processes (red boxes).
  (b) Execution time for command handoff in each control cycle. Largest time measured was 57.648\,$\mu$s (within WCET 98.299\,$\mu$s, Figure \ref{fig:jitter}a). 
  (c) Release Jitter for each control cycle. Maximum jitter measured was 0.703\,$\mu$s (within MRJ 3.110\,$\mu$s, Figure \ref{fig:jitter}b). 
  After compilation is complete for each new controller, they are hot-swapped into the Spine at approximately $t=5, 9, 16$s (blue arrows)
  (d) The robot's tracking performance across four hot-swapped real-time controllers, at 1000Hz. No real-time deadlines were missed. Note the initial startup delay for $t<2.2$s to initialize LITHE. At $t=21.5$s, we simulate a brain crash, but LITHE continues to hold the load smoothly against gravity, preventing mechanical collapse. }
  \label{fig:decentralized}
\end{figure}

The experiment proceeded in four automated phases (Fig. \ref{fig:decentralized}):

\begin{itemize}
    \item \textbf{Baseline:} The robot tracked a 0.5 Hz sine wave using a detuned PD controller. High tracking error was expected due to the lack of gravity compensation.
    \item \textbf{Probing:} The Agent commanded a System Identification routine. It generated and hot-swapped two probing controllers that held constant torque and measured steady-state deflection. The Agent then estimated the gravitational constant ($\widehat{mgl}$) from this.
    \item \textbf{Evolution:} Using the estimated parameters, the Agent wrote a new C++ control law incorporating feed-forward gravity compensation ($\tau_{ff} = \widehat{mgl} \sin \theta$) using the newly estimated parameters. Note that $\theta$ is continuously updated position feedback within the active (1 kHz) real-time loop. This code was transmitted to LITHE, compiled, and hot-swapped.
    \item \textbf{Fault Tolerance Test:} To simulate a catastrophic failure of the high-level software, we manually froze the Brain process (CPU 2) for 1.5 seconds while the arm was under maximum gravity load.
\end{itemize}

\subsubsection{Results} 
Figure \ref{fig:decentralized} visualizes the system's evolution. The duration of each agent-supervised phase (Fig. \ref{fig:decentralized}a) was not deterministic, because it relied on highly variable computation time of the LLM robot client. Crucially, throughout all hot-swaps and network delays, LITHE maintained functional real-time operation with no missed deadlines.

\textbf{Performance Improvement:} 
The Baseline phase shows significant tracking error (Fig. \ref{fig:decentralized}d), with root mean squared error (RMSE) of 71.7 deg for $t \in [2, 5]$. The transition to the Evolved Controller ($t \approx 16$s) yields an immediate reduction in error, with RMSE of 43.0 deg for $t \in [16, 21.5]$, and 18.9 deg for $t \in [23.5, 28.5]$ after the fault tolerance test. A minor residual error remains, indicating a slight mismatch in the LLM's estimated $\widehat{mgl}$. Still, integration of the feed-forward term demonstrates that the Agent successfully synthesized and deployed a context-aware real-time control law.

\textbf{Jitter Stability:} 
During compilation and hot-swap events (Fig. \ref{fig:decentralized}b, \ref{fig:decentralized}c), execution jitter on the Spine peaked at 57.6\,$\mu$s (Fig. \ref{fig:decentralized}c), and release jitter remained below 0.8\,$\mu$s (Fig. \ref{fig:decentralized}d). This confirms that the heavy system calls associated with dynamic linking were successfully isolated to the Loader thread (CPU 0), leaving the real-time control loop unaffected.

\textbf{Fault Tolerance:} 
At $t \in [21.5, 23]$, the Python Brain was frozen (Fig. \ref{fig:decentralized}d). Despite the total loss of reference commands, the Spine continued to execute its real-time control law (gravity compensation) using its last received command. The robot maintained its position smoothly (Fig. \ref{fig:decentralized}b, yellow trace), because the isolated controller continuously corrected for external perturbation (gravity) in real-time, preventing mechanical collapse. LITHE successfully decoupled low-level safety from high-level software instability.

\section{DISCUSSION}
\label{sec:discussion}

\subsection{Characterizing User-Space Real-Time}
The central premise of LITHE is that modern commodity CPUs are sufficiently fast to brute-force 1000 Hz real-time performance in user space, if the OS scheduler is forcibly excluded. Our results support this. Release jitter (Fig. \ref{fig:jitter}b) remained consistently $<3.11\,\mu$s because the Spine's spin-wait approach eliminates wake-up latency inherent to OS sleep calls. Notably, the NumPy thread sprawl failed to penetrate the Spine's isolation, confirming that the critical real-time core is effectively invisible to the Linux scheduler.

The distribution of this jitter (Fig. \ref{fig:jitter}) is irregular and non-Gaussian with positive skew, unlike typical electrical noise. This ``heavy-tailed'' behavior likely stems from irreducible hardware resource contention, such as L3 cache misses or memory bus arbitration, which persist even when the CPU core itself is isolated.

This contention was visible during the compilation phases of our supervised evolution demonstration. We observed a jitter density increase (Fig. \ref{fig:decentralized}b, \ref{fig:decentralized}c) prior to controller swaps. This likely corresponds to the compiler (\texttt{gcc}) on the Housekeeping core competing for memory bandwidth with the real-time Spine. Significantly, this contention never pushed beyond the critical 1 ms deadline (max 57.6\,$\mu$s $<$ WCET 98.299\,$\mu$s). This validates that for 1 kHz applications, the ``User-Space Real-Time'' approach provides a functional margin of safety, even if it lacks the formal mathematical guarantees of a verified real-time operating system.

\subsection{Integration with Decentralized Agents} 
Real-time robotics typically requires immutable, pre-compiled machine code running on a processor. LITHE leverages this compiled efficiency while simultaneously enabling decentralized and best effort processes to evolve real-time logic on-the-fly. In our experiment, the Probing and Evolution phases (Fig. \ref{fig:decentralized}) demonstrated this flexibility. The external Agent did not simply tweak parameters or generate a trajectory; it fundamentally altered the real-time control structure to inject probing torques and evolve the controller. While we utilized an external LLM to demonstrate the extreme case of high-latency supervision, the architecture is agnostic to intelligence source; a local Reinforcement Learning (RL) policy running on the RPI's Python Brain might  update the control law with lower latency and complexity. In this ML paradigm, it remains an area of active research to implement appropriate safety and verification bounds on the model's output. Nonetheless, LITHE's capability allows the ``Thinking'' (Python/Brain/Model) and the ``Acting'' (C++/Spine) aspects of a robot to operate on completely decoupled timescales, resolving the immutable bottleneck that hampers evolutionary real-time control.

\subsection{Limitations and Boundaries} 
\label{disc:limits}
While LITHE offers an accessible scaffold for dynamic machine-learning architecture under real-time, it operates within specific physical and architectural boundaries.

\textbf{Bandwidth vs. Compute:} Our 1 kHz loop rate was constrained by the round-trip latency of the CAN-FD bus, not the Raspberry Pi's CPU. As noted by Pieper \cite{OptimizingMoteusCommand}, the \textit{pi3hat} can theoretically sustain 1 kHz with four devices on a single bus (it has five independent buses), and we have separately validated LITHE in ``Frankenstein'' configurations mixing Moteus CAN-FD and Teensy-based CAN gateways. Still, scaling to complex multi-device topologies requires careful bus-load balancing to ensure total bus round-trip time remains within the 1 ms budget. Moreover, while our pipelined execution maximizes the Spine's machine cycles, a complex or unoptimized control law algorithm may generate computation that exceeds the 1ms deadline without a faster CPU. Additionally, while bandwidth was our main bottleneck, LITHE's full core utilization strategy (with busy-wait spinlocks) increases power consumption. This may lead to thermal throttling and compute performance degradation, which necessitate active thermal management strategies.

\textbf{State Variable Continuity:} 
Hot-swapping controllers with disparate internal states (e.g., integral errors or filtered velocities) risks discontinuity. To ensure smooth transitions, we map these state variables to the persistent shared memory block, outside the transient controller object. Still, evolving a controller that requires entirely new state definitions remains a challenge. We mitigate this using a ring buffer IPC (rather than a single-snapshot IPC), to enable nonzero-valued initialization of new states. Still, because the ring buffer size must be pre-allocated, the smoothness of transitioning between any arbitrarily complex evolved controller is inherently bounded.

\textbf{The Compilation Bottleneck:} The Hot-Swap mechanism eliminates \textit{deployment} friction, but \textit{generation} friction remains. In figures \ref{fig:decentralized}c and \ref{fig:decentralized}d, \texttt{gcc} compilation on the Brain core took $\approx$2.5 seconds. While this does not interrupt the robot, it limits the rate of evolutionary steps. Future versions of LITHE could mitigate this latency with Tiny C Compiler (TCC) or Just-In-Time (JIT) engines (e.g., ClangJIT) to bypass filesystem access and emit machine code directly into memory. Alternatively, the compilation bottleneck can be minimized for fixed-structure controllers by exposing gains as tunable parameters in shared memory, and restricting full recompilation to cases where the control \textit{structure} itself must change (e.g., from PID to a linear quadratic regulator).

\textbf{System-Level Safety:} Finally, we emphasize that ``Functional Real-Time'' is not ``Hard Real-Time.'' LITHE provides process-level safety (surviving a Python crash), but the Linux kernel must remain active. For safety-critical human-robot interaction, this architecture should always be augmented with a hardware-level watchdog (e.g., on the STM32 layer) to clamp motor outputs in the event of total kernel panic. Furthermore, in the paradigm of ML-evolved control, theoretical stability guarantees remain an open challenge. Where control theory is unvalidated, safety must be enforced via strict hardware-level limits on torque and velocity.

\subsection{Generalizability and System Integration}
LITHE makes real-time programming more accessible and flexible, because it does not require advanced knowledge in \texttt{PREEMPT\_RT} kernel patching, which can be brittle and fickle with system or driver updates. Given that LITHE implements on vanilla Raspberry Pi OS Lite \textbf{without} \texttt{PREEMPT\_RT}, it is easily deployable on other single-board compute (SBC) running vanilla Linux using similar boot parameters (see project website, e.g., \texttt{isolcpus=1-3}). This means that any aforementioned bandwidth or compute bottlenecks can be addressed with more capable compute hardware (e.g., with RPI 5 or NVIDIA Jetson). Furthermore, LITHE architecture and \texttt{PREEMPT\_RT} are not mutually exclusive because LITHE relies on user-space \textit{evasion} instead of kernel-space \textit{preemption}, so potential jitter reduction in Figure \ref{fig:jitter}c and \ref{fig:jitter}d may be possible with a well-constructed \texttt{PREEMPT\_RT} transport thread from the \textit{pi3hat} library. 

LITHE can also be flexibly integrated into preexisting robot systems. In our demonstration with decentralized agents, we turned the Brain's robot application into a downstream node from an agentic robot client over TCP/IP (Fig. \ref{fig:architecture}). In a similar fashion, the Brain can interface with other robot systems, such as ROS-2, by subscribing to interface definition language (IDL) topics that create commands and trajectories for the Spine. LITHE thus has value as a generalizable and accessible real-time prototyping tool. 

\subsection{Future Directions: Morphological Evolution for Wearable Soft Robotics}
The LITHE architecture opens a pathway for accessible high-performance research, particularly in our domain of soft robotics and wearable devices. In wearable human-robot interaction, the timescales of control are bifurcated. At the millisecond scale, stiff real-time control is required to ensure safe, compliant reaction forces that minimize damage to soft tissue across a spectrum of activities. At the diurnal timescale (minutes to hours), the controller must evolve to account for the viscoelasticity, creep, and shape-memory effects that are inherent to physical interaction between hardware and biology. We envision LITHE as a low-cost platform for bridging these timescales. Morphological evolution of the controller can be implemented with hot-swapped control laws to fundamentally adapt \textit{with} the soft tissue, across different activities and tasks, without interrupting stiff real-time operation. We are avidly exploring this area of research.

\subsection{Towards General Embodied Intelligence}
Ultimately, safe embodied intelligence requires reflexes that are faster than its thoughts. Consider a service robot navigating a crowded convention hall: it may bump into a human before its high-level vision system processes the event \cite{wangBrainReflexEmergency2026}. If it relies on a Python-based planner to react, it may be too late. LITHE addresses this by ensuring that spinal reflexes remain deterministic and robust, regardless of the cognitive load or latency of the higher-level Brain. Concurrently, LITHE allows for these spinal reflexes to be updated and changed as needed for general purpose robots, whether it be for safe human-interacting environments or a precise industrial manipulation. By democratizing these capabilities on low-cost commodity hardware, we aim to accelerate the transition from rigid, pre-programmed automation to adaptive, resilient, and continuously evolving intelligent systems.

\section{CONCLUSION}
\label{sec:conclusion}

In this work, we introduced LITHE, a lightweight architecture that enables functional real-time robotic control on low-cost commodity Linux hardware without the need for specialized RTOS kernels or proprietary RCP systems. By strictly isolating the  ``Spinal'' control loop from the ``Cerebral'' high-level logic, we demonstrated that a Raspberry Pi 4B can sustain 1 kHz actuation with WCET $<100\,\mu$s and MRJ $<4\,\mu$s, even under heavy computational load.

Furthermore, we validated the system's ability to hot-swap control laws at runtime, overcoming the immutability bottleneck and enabling a new class of ``evolving real-time'' robots. We demonstrated this by coupling LITHE to a decentralized supervisory agent, which successfully performed system identification and deployed a gravity compensation controller to the moving robot without interrupting its real-time operation. LITHE lowers the barrier to entry for advanced evolutionary real-time control, providing a robust foundation for the next generation of embodied intelligence that is both agile in reaction and flexible in adaptation.

\section*{ACKNOWLEDGMENT}
We thank 
Richard Lin
and members of the 
UCLA Robotics
community who provided invaluable feedback and insight into this work.

\subsection{Declaration on Use of AI Tools}
Gemini 2.5 Pro, 3.0 Pro, and 3.1 Pro were used via the web interface to survey literature by parsing through open-source code repositories, conceptualize parts of LITHE system architecture, generate code snippets (edited manually in Visual Studio Code), and debug problems in the software. qwen2.5-coder-7b was deployed with our custom agent for the experiment on supervised controller evolution.


\bibliographystyle{IEEEtran}
\bibliography{references}


\end{document}